\documentclass{article}
\usepackage[nonatbib, final]{nips_2018}
\usepackage{rotating}
\usepackage{bookmark}
\usepackage{natbib}
\usepackage{algorithm}

\usepackage{amsmath}
\usepackage{bm}
\usepackage{graphicx}
\usepackage{float}
 \usepackage{algpseudocode} 
\usepackage[utf8]{inputenc}
\usepackage[T1]{fontenc}    
\usepackage{url}          
\usepackage[titletoc]{appendix}
\usepackage{booktabs}       
\usepackage{amsfonts}       
\usepackage{nicefrac}       
\usepackage{microtype} 
\usepackage{xcolor}

\newcommand{\overbar}[1]{\mkern 1.5mu\overline{\mkern-1.5mu#1\mkern-1.5mu}\mkern 1.5mu}

\title{A Fast and Greedy Subset-of-Data (SoD) Scheme for Sparsification in Gaussian processes}

\author{
Vidhi Lalchand  \qquad \qquad  A. C. Faul \\
University of Cambridge, Cambridge, UK\\
The Alan Turing Institute, London, UK\\
\texttt{\{vr308, acf22\}@cam.ac.uk}
}

\begin{document}
\maketitle
\begin{abstract}
In their standard form Gaussian processes (GPs) provide a powerful non-parametric framework for regression and classificaton tasks. Their one limiting property is their $\mathcal{O}(N^{3})$ scaling where $N$ is the number of training data points. In this paper we present a framework for GP training with sequential selection of training data points using an intuitive selection metric. The greedy forward selection strategy is devised to target two factors - regions of high predictive uncertainty and underfit. Under this technique the complexity of GP training is reduced to $\mathcal{O}(M^{3})$ where $(M \ll N)$ if $M$ data points (out of $N$) are eventually selected. The sequential nature of the algorithm circumvents the need to invert the covariance matrix of dimension $N \times N$ and enables the use of favourable matrix inverse update identities. We outline the algorithm and sequential updates to the posterior mean and variance. We demonstrate our method on selected one dimensional functions and show that the loss in accuracy due to using a subset of data points is marginal compared to the computational gains.
\end{abstract}

\section{Introduction}

Gaussian processes are nonparametric tools, allowing the complexity of the model to grow as more data is observed. Another attractive feature of GPs is the behaviour of the predictive variance which is naturally higher in regions away from the training data. This is intuitive as in regions where there is no training data there is higher uncertainty about the interpolating function. The application of GPs to the regression task involves the computation of a matrix inverse, this leads to the $\mathcal{O}(N^{3})$ scaling. Further, $\mathcal{O}(N^{2})$ space is required to store a dense covariance matrix in memory. 

There has been significant interest in finding sparse approximations to the full Gaussian process in order to speed up training and prediction times to $\mathcal{O}(NM^{2})$ where $M$ is the size of an auxiliary set, typically a subset of the training data. \citet{quinonero2005unifying} provides a unifying summary of sparse approximations. A common theme in some of the earlier sparse methods involved developing a low-rank approximation to the covariance matrix, also called the Nystr\"{o}m approximation \citep{smola2000sparse, seeger2003fast}.  In these schemes the full covariance matrix of size $N$ is replaced by the  Nystr\"{o}m approximation requiring the inverse of a smaller matrix involving $M$ data points. These class of methods are called \emph{projected} process approximation schemes in \citet{quinonero2005unifying}. Most of the recent innovations in this field are driven by the variational approach in \citet{titsias2009variational}. 

In this short paper we focus on a greedy approximation scheme which results in a sequential construction of a subset of size $M$ from the $N$ training data points. The selection metric used to rank points for selection can be evaluated with low computational overhead. It might be worth noting that the task of selecting the active set in the context of regression is a general idea that can be coupled with different projected process approximation schemes to generate new methodologies. 

\section{GP Regression (GPR)}

A GP is a collection of random variables $\{f(\mathbf{x})| \mathbf{x} \in X \}$, any finite number of which have a joint Gaussian distribution. A GP is fully specifed by the mean function $\mu(\mathbf{x})$ and covariance function $k(\mathbf{x}, \mathbf{x'})$ which are user defined, the covarinace function typically depends on a set of hyperparameters $\bm{\theta}$. GPs can be used to define a distribution over functions $f(\mathbf{x}) \sim \mathcal{GP}(\mu, k)$ as they can be viewed as a collection of random variables,  this means that any finite collection of function values $[f(\mathbf{x_{1}}), \ldots, f(\mathbf{x_{N}})]$ have a joint Gaussian distribution. 

\begin{align}
[f(\mathbf{x_{1}}), \ldots, f(\mathbf{x_{N}})] \sim \mathcal{N}(\bm{\mu},K)
\end{align}

where $\bm{\mu}$ is the $N \times 1$ vector $\mu_{i} = \mu(\mathbf{x_{i}})$ and $K$ is the $N \times N$ covariance matrix with $K_{ij} = k(\mathbf{x_{i}, \mathbf{x_{j}}})$. 

Our training dataset consists of $N$ pairs of data $(\mathbf{x_{i}}, y_{i})_{i=1}^{N}$  where $y_{i}$ are noisy realisations of some latent function $f$ with Gaussian noise $y_{i} = f(\mathbf{x}_{i}) + \epsilon_{i}$, $\epsilon_{i} \in \mathcal{N}(0, \sigma^{2})$. Let $X$, $\mathbf{y}$ denote the training inputs and noisy targets and $\mathbf{f}$ denote the vector of underlying latent function values. The liklihood of the data $\mathbf{y}|\mathbf{f} \sim  \mathcal{N}(f, \sigma^{2}I)$ and the prior $\mathbf{f} \sim \mathcal{N}(0, K)$ give the joint probability model $p(\mathbf{f}, \mathbf{y}) = p(\mathbf{y}|\mathbf{f})p(\mathbf{f})$. The predictive distribution at a set of test inputs $X_{*}$ is given in closed form using properties of conditional Gaussians, 

\begin{align}
\begin{split}
\mathbf{f}_{*}|\mathbf{y}, X, X_{*}, \bm{\theta}, \sigma^{2} &\sim \mathcal{N}(\overbar{\mathbf{f}_{*}}, cov(\mathbf{f}_{*})) \\
\overbar{\mathbf{f}_{*}} &= K_{*}(K + \sigma^2I)^{-1}\mathbf{y} \\
cov(\mathbf{f}_{*}) &= K_{**} -  K_{*}(K + \sigma^{2}I)^{-1}K_{*}^{T}
\label{pred}
\end{split}
\end{align}

where $K_{**}$ denotes the covariance matrix evaluated between the test inputs $X_{*}$ and $K_{*}$ denotes the covariance matrix evaluated between the test inputs $X_{*}$ and training inputs $X$, if there are $N_{*}$ test inputs the covariance matrix $K_{**}$ is of size $N_{*} \times N_{*}$ and $K_{*}$ is of size $N_{*} \times N$. The hyperparameters along with the noise variance $(\bm{\theta}, \sigma^{2})$ are inferred through optimisation of the log marginal likelihood given by $\int p(\mathbf{y}|\mathbf{f})p(\mathbf{f})d\mathbf{f}$ which can be analytically derived through marginalising $\mathbf{f}.$ 


\section{Greedy framework for Gaussian Process Regression}
\label{pfgpr}

The aim is to select a smaller informative subset $\mathbf{u} \in \mathbf{y}$ (the \textit{active} set) which play an active role in the inference. All other training points $\mathbf{y} \backslash \mathbf{u}$ belong to the \textit{remainder} set. The active set  is constructed incrementally, at each iteration exactly one training data point is selected. Following the notation from \citet{seeger2003fast} let $I$ denote indices of the active set and $R = \{1, \ldots, N\} \backslash I$ denote the indices of remainder points. Training happens in stages we index by $t$. Hence, $I_{t}$ and $R_{t}$ denote the index sets at stage $t$ of training.

We denote the active set $\mathbf{u}$ by $\mathbf{y}(I_{t})$ from here on to clearly incorporate the stage of training. Similarly, the remainder set is denoted as $\mathbf{y}(R_{t})$. The active and remainder input locations are denoted as $X(I_{t})$ and $X(R_{t})$ respectively. For the purposes of testing we have a hold-out set with $N_{*}$ test inputs $X_{*}$ and targets $\mathbf{y_{*}}$. 

\begin{table}[H]
\begin{center}
\begin{tabular}{|l|c|c|c|c|}
\hline
\multicolumn{5}{|c|}{\bfseries At stage $t$} \\
\hline
 & Index set & Inputs & Targets & Size \\
 \hline
Active points & $I_{t}$  & $X(I_{t})$  & $\mathbf{y}(I_{t})$   & $t$ \\
Remainder points &  $R_{t} $  &  $X(R_{t})$  & $\mathbf{y}(R_{t})$ & $N-t$ \\
\hline
\end{tabular}
\caption{Notation used in stagewise training}
\end{center}
\end{table}
\vspace{-5mm}
At stage $t$ the active set has exactly $t$ data points as at each stage exactly one point is added to the active set. We have a fixed set of $N$ training pairs, the active set grows in size as more points are added to it and the remainder set shrinks in size as points are removed from it. Essentially, we start with all the training data points in the remainder set and points move from the remainder set to the active set in each iteration based on a selection criteria. 

\begin{itemize}
\setlength\itemsep{-0.2em}
\item $K_{{I}_{t}} = K(X(I_{t}), X(I_{t}))$ denotes the $t \times t$ covariance matrix computed between the active inputs at the $t^{th}$ stage of training.
\item $K_{\backslash I_{t}} = K(X(R_{t}), X(I_{t}))$ denotes the $(N-t)\times t$ covariance matrix computed between the $t$ active inputs selected so far and the $(N-t)$ remainder inputs in $R_{t}$.
\item $K_{R_{t}} =  K(X(R_{t}), X(R_{t}))$  denotes the $(N-t) \times (N-t)$ covariance matrix computed between the remainder inputs at stage $t$.
\item $\bm{\mu}_{t}$ and $\Sigma_{t}$ denote the predictive posterior mean and covariance computed at stage $t$ for the remainder inputs $X(R_{t})$. 
\end{itemize}

The algorithm starts with a single training point $\mathbf{y}(I_{1}) = [u_{1}]$ in the active set which is selected at random from $\mathbf{y}$,  the predictive posterior mean and covariance denoted by $\bm{\mu_{1}}$ and $\Sigma_{1}$ at stage 1 are computed by conditioning on the active set (of 1 point) while the goodness of fit is assessed by predicting on the remainder inputs ($N-1$ points). In short, we predict at each stage the mean and covariance of the remainder inputs $X(R_{t})$ and compare them to the true remainder targets $\mathbf{y}(R_{t})$. The mean squared error computed between the true remainder targets $\mathbf{y}(R_{t}) = \{ r_{i} | i \in R_{t} \}$ and the predicted mean $\bm{\mu}_{t}$ provides the basis for convergence. If the decrease in $|\bm{\mu}_{t} - \mathbf{y}(R_{t})|_{2} = \sum_{i=1}^{N-t}(\mu_{i} - r_{i})^{2}$ is under a threshold, we terminate.

The main reason for this stagewise iterative training approach is two-fold:

\begin{enumerate}
\item The selection criteria for the active training target at each stage is tied to the predictive posterior mean and variance computed on the remainder inputs. 
\item Since the active set is grown one point at a time, we can take advantage of favourable matrix inverse update identities in order to update the predictive posterior mean and variance at each stage (see \ref{fsu} for a detailed discussion).
\end{enumerate}

At stage t, 
\begin{align}
\mathbf{y}(R_{t})|&\mathbf{y}(I_{t}), X(I_{t}), X(R_{t})  \sim  \mathcal{N}(\bm{\mu_{t}}, \Sigma_{t})  \\
\bm{\mu}_{t} &= K_{\backslash I_{t}}(K_{I_{t}} + \sigma^{2}_{n}\mathbb{I})^{-1}\mathbf{y}(I_{t}) \nonumber \\
\Sigma_{t} &= K_{R_{t}} - K_{\backslash I_{t}}(K_{I_{t}} + \sigma^{2}_{n}\mathbb{I})^{-1}K_{\backslash I_{t}}^{T}  
\label{predr}
\end{align}

The above equations reflect the predictive posterior mean and covariance for a full GP introduced  in eq. \ref{pred} where the active set $\mathbf{y}(I_{t})$ plays the role of the target vector $\mathbf{y}$. The hyperparameters $(\bm{\theta}, \sigma^{2})$ are kept fixed during the greedy training. They are estimated by optimising the marginal likelihood on a random subset of the training data (see section \ref{sh} for a discussion).

\subsection{The Algorithm}

Below we give the general algorithm for active set selection using a general selection metric we denote as $\Delta$. 

\begin{algorithm}[H]
\hspace*{25pt}
\caption{Greedy framework for GPR}
\begin{algorithmic}
\State \textcolor{blue}{Initialisation:} Pick a random target $u_{1} \in \mathbf{y}$. 
\State \textcolor{blue}{Convergence Condition:} $(RMSE_{t-1} - RMSE_{t}) < \delta$ calculated on the remainder set $ \bm{y}(R_{t}) = \bm{y}\backslash \bm{y}(I_{t})$.
\State \textbf{for} each stage $t$: 
\State \qquad \textcolor{blue}{GP Train} on $(\bm{X}(I_{t}), \bm{y}(I_{t}))$.
\State \qquad \textcolor{blue}{GP Predict} on $\bm{X}(R_{t})$.
\State \qquad Update posterior mean $\mu_{t}$ and covariance $\Sigma_{t}$. (see section \ref{fsu} for sequential update rules.)
\State \qquad Compute $\Delta_{i} \forall i \in R_{t}$ 
\State \qquad Select $i$ where $i = argmax_{i \in R_{t}} \Delta_{i}$ 
\State \qquad $I_{t+1} \leftarrow I_{t} \cup \{i\}, R_{t+1} \leftarrow R_{t} \backslash \{i\}$ 
\State \qquad \textbf{If} \textcolor{blue}{convergence} is \textbf{true}:
\State \qquad \qquad break;
\State \textbf{end} 
return $I_{T}, R_{T}$  
\end{algorithmic}
\end{algorithm}

The algorithm advances in stages by selecting the next active point from the remainder set as the maximiser of the selection criteria $\Delta$ given by the additive term:
\vspace{-2mm}
\begin{equation*}
argmax_{u \in \bm{y}(R_{t})} \Delta = \overbrace{\sqrt{diag(\Sigma_{t})}}^{uncertainty} + \overbrace{|(\bm{\mu}_{t} - \bm{y}(R_{t}))|}^{underfit}
\end{equation*} 

The component terms in $\Delta$ address the two-fold objective of targeting regions of high uncertianty captured by the term  $\sqrt{diag(\Sigma_{t})}$ which is the posterior predicted standard deviation at stage $t$ and underfit captured by the error term $|(\bm{\mu}_{t} - \bm{y}(R_{t}))|$ which denotes the deviation of the remainder targets from the posterior predicted mean. 

\begin{figure}
\begin{center}

\includegraphics[scale=0.55]{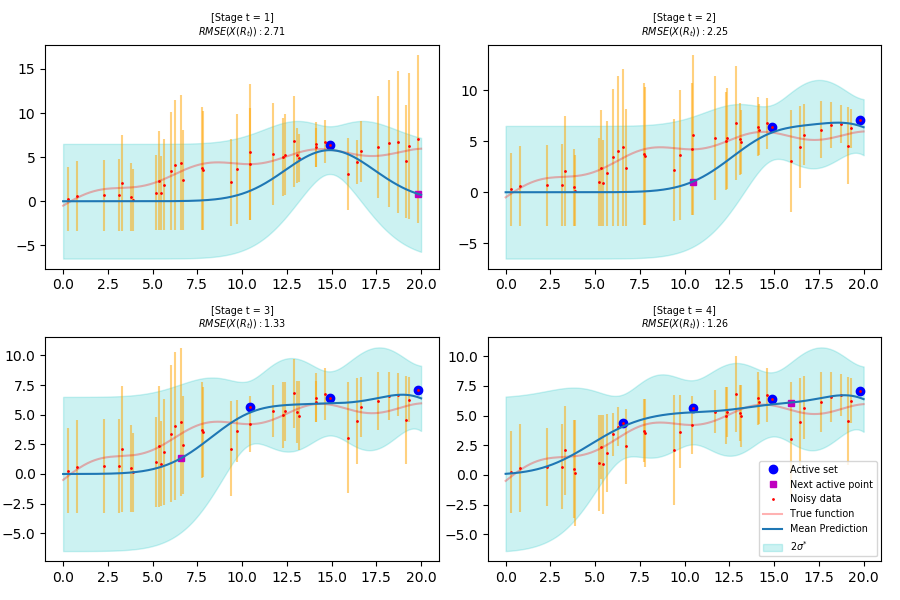}
\caption{A demonstration of the greedy GP training (stages 1-4). The orange vertical lines denote the $\Delta$ selection metric. The algorithm terminates after 13 iterations; here we only depict the first 4 iterations owing to lack of space.}
\label{demo}
\vspace{-7mm}
\end{center}
\end{figure}

Note that both components of the addition are vectors of size $(N-t)$ as they are based on the remainder set. The metric $\Delta$ can be evaluated in $\mathcal{O}(1)$ as $\Sigma_{t}$ and $\bm{\mu}_{t}$ are obtained directly from the training step. A visual depiction of the evolution of greedy training for the function $x\sin{x}$ is shown in fig. \ref{demo}.  The computational complexity of the full greedy training algorithm is given in section \ref{cppt}. 

\vspace{-2mm}

\subsection{Experiments}
\label{ne}

We trained a GP using the greedy training approach by sampling noisy \footnote{In order to have a systematic comparison the noise level for all the experiments was identical.} values from a host of $1d$ functions; we then predict on a hold out unseen test set $X_{*}$. We report the generalisation error (RMSE) under three \footnote{The random subset scheme used targets sampled uniformly from the training data and were ensured to be the same size as that for the Greedy GP.}  training schemes. The squared exponential (SE) kernel was used in all the three schemes (see section \ref{sh}). 

\begin{table}[H]
\centering
\begin{tabular}{l|c|c|c|c}
Function $\backslash$ Data & Full GP & Random & Greedy GP &  $\%$ of full dataset \\
\hline \hline 
$x^{2}sin(x)$  &  32.24 & 91.62 & 39.29 &  $22\%$ \\
$xsin(x)$ & 2.36 & 5.95 & 2.82 & 18$\%$\\
$0.5sin(x) + 0.5x - 0.02(x-5)^{2}$ & 1.14 & 2.17 & 1.96 & $31\%$\\
\end{tabular}
\caption{RMSE on Test data}
\end{table}

\subsubsection*{Acknowledgments}

This work was supported by the Alan Turing Institute through the Doctoral Studentship for International Students.


\bibliographystyle{authordate1}
\bibliography{ref}

\appendix

\section{Appendix}

\subsection{Greedy updates}
\label{fsu}

In this section, we discuss in detail the stagewise updates to the posterior predicted mean and covariance given in eq. \ref{predr}.
\subsubsection{Mean}

\begin{align*}
\textrm{At stage t: } \mu_{t} &= K_{\backslash I_{t}}(K_{I_{t}} + \sigma^{2}\mathbb{I})^{-1}\bm{y}(I_{t})\\
\downarrow \\
\textrm{At stage t+1: } \mu_{t+1} &= K_{\backslash I_{t+1}}(K_{I_{t+1}} + \sigma^{2}\mathbb{I})^{-1}\bm{y}(I_{t+1})
\end{align*}

Notice that the mean update from stage $t \rightarrow t+1$ involves updating two matrices. \\

First, 
\begin{equation*}
 K_{\backslash I_{t}} \rightarrow  K_{\backslash I_{t+1}}
\end{equation*}

In this update we evolve a $(N-t) \times t$ matrix  into a $(N - (t+1)) \times (t+1)$ . We are dropping a row and adding a column. The newly added column contains the covariances computed between the newly selected active point, say $s$ and all the $(N - (t+1))$ points in the remainder set $(k(s, r_{i}) | i  \in R_{t+1})$. If we assume that the full covariance matrix $K(X,X)$ is computed at the start then this update just requires selecting the corresponding entries from the matrix $K(X,X)$. The complexity of this operation is just $\mathcal{O}(1)$.\\

Second, 

\begin{equation*}
K_{I_{t}} \rightarrow K_{I_{t+1}}
\end{equation*}

This is the covariance matrix computed on the active points, it grows by 1 row and 1 column in each stage. 

\begin{figure}[H]
\includegraphics[scale=0.5]{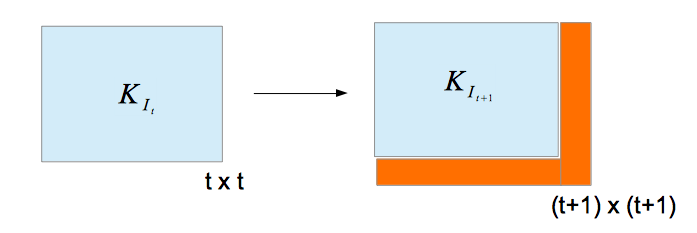}
\end{figure}

Further, its inverse is required in each stage. We make use of block inversion to update the inverse. The complexity of this operation is quadratic per iteration, this is shown in section \ref{cppt}.

\subsubsection{Covariance}

\begin{align*}
\textrm{At stage t: } \Sigma_{t} &= K_{R_{t}} - K_{\backslash I_{t}}(K_{I_{t}} + \sigma^{2}\mathbb{I}_{t})^{-1}K_{\backslash I_{t}}^{T}\\
\downarrow \\
\textrm{At stage t+1: } \Sigma_{t+1} &= K_{R_{t+1}} - K_{\backslash I_{t+1}}(K_{I_{t+1}} + \sigma^{2}\mathbb{I}_{t})^{-1}K_{\backslash I_{t+1}}^{T}\\
\end{align*}

The updates of $K_{\backslash I_{t}} \rightarrow  K_{\backslash I_{t+1}}$ and $K_{I_{t}} \rightarrow K_{I_{t+1}}$ were already discussed in the previous section. The only new matrix update involved here is,

\begin{equation*}
K_{R_{t}} \rightarrow K_{R_{t+1}}
\end{equation*}

Since, $K_{R_{t}}$ is the covariance matrix computed on the remainder inputs and the size of the remainder inputs shrinks as the training progresses, $K_{R_{t}} \rightarrow K_{R_{t+1}}$ involves dropping a row and a column as the matrix shrinks from size $(N-t) \times (N-t) $ to $(N - (t+1)) \times (N - (t+1))$.

\subsection{Complexity of Greedy training}
\label{cppt}

The cost of computing an updated inverse for a square matrix of size $M$  grown by 1 row and 1 column is quadratic $\mathcal{O}(M^{2})$ if we use block inversion (Schur complement) and assuming we know the inverse of the matrix of size $M$. 
In the greedy GP framework, in each stage the covariance matrix computed on active points grows by 1 row and 1 column. If we end up with an active set of $M$ points after $M$ stages of training we have conducted the matrix update operation $M$ times and the cost each time is quadratic in the dimension of the matrix we are updating. This give us the following computational cost in terms of operations. 

\begin{equation}
\sum_{i=1}^{M}i^{2} = 1^{2} + 2^{2} + \ldots + M^{2} = \dfrac{M(M+1)(2M+1)}{6} = \mathcal{O}(M^{3})
\end{equation}

Hence, while the order of complexity is quadratic per iteration, the overall complexity is $\mathcal{O}(M^{3})$ if $M$ is the size of the final active set.  It is important to note that if the inverse if calculated directly in each iteration, the complexity is $\mathcal{O}(M^{4})$; hence, the update with the Schur complement is essential. \\

The table below highlights the computational complexity of the full GP and the greedy approach to training. Note that, we still need to compute and store the full covariance matrix in order to speed up the matrix updates in the greedy GP approach. 

\begin{center}
\begin{tabular}{|l|c|c|}
 Task & Full GP & Greedy GP \\
\hline
Training & $\mathcal{O}(N^{3})$ & $\mathcal{O}(M^{3})$ \\
Prediction & $\mathcal{O}(N^{2})$ & $\mathcal{O}(M^{2}N)$\\
Storage (for K) &  $\mathcal{O}(N^{2})$ &  $\mathcal{O}(N^{2})$
\end{tabular}
\end{center}

\subsection{Selection of hyperparameters}
\label{sh}

The \textit{squared exponential}(SE) kernel is defined as, 

\begin{equation}
k(\bm{x}_{i}, \bm{x}_{j}) = \sigma^{2}_{f}\exp{\Big(-\dfrac{(\bm{x}_{i}-\bm{x}_{j})}{2l^{2}}}\Big) + \delta_{ij}\sigma^{2}_{n}
\label{se}
\end{equation}

where $\theta = ( \sigma^{2}_{f}, l,  \sigma^{2}_{n})$ is the set of hyperparameters, comprising the signal variance $\sigma_{f}^{2} > 0$ which controls the variation of function values from their mean, the lengthscale $l > 0$ which controls how smooth a function is and $\sigma_{n}^{2} >= 0$ is the noise variance which allows for noise to be present in the data. The noise variance applies only when $i = j$. In the noiseless case, we just drop the additive noise variance term.\\ 

\begin{figure}[H]
\centering
\includegraphics[scale=0.6]{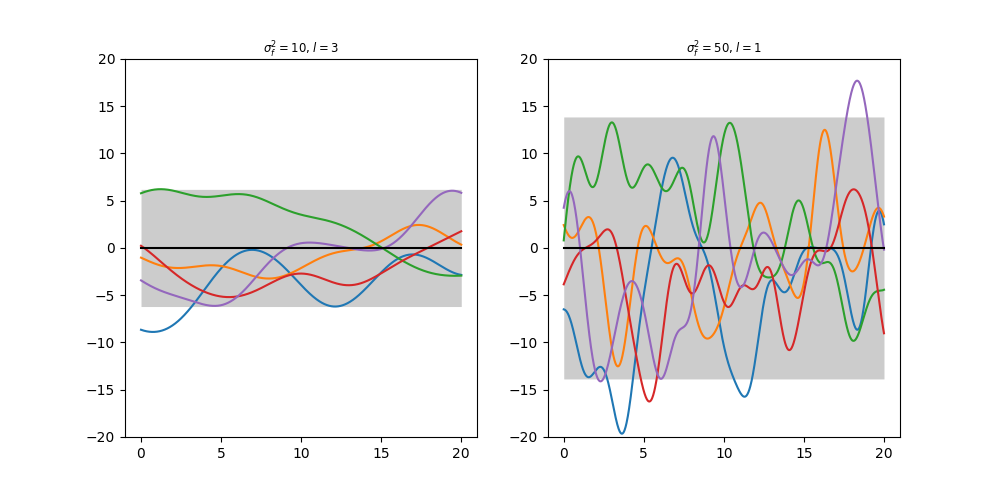}
\caption{Samples from GP prior with SE covariance function and 95\% confidence intervals $(\pm 1.96\sigma_{f}$)}
\label{cont}
\end{figure}

The hyperparamters for the SE kernel used in the covariance matrix $\theta=(\sigma^{2}_{f}, l,  \sigma^{2}_{n})$ are pre-selected through optimization of the log marginal likelihood (LML) using a random subset of the training data in a pre-processing step. During the running of the algorithm, the hyperparameters remain fixed. In this paper, we mainly focus our efforts at providing a framework for selecting active targets and inputs from the training points while simultaneously training the GP. A framework that weaves together the hyperparameter selection and active set selection in the context of greedy training of GPs is being researched. Preliminary experiments where we varied the hyperparameters during stagewise training using marginal likelihood optimisation lead to instability.  The authors of \citet{snelson2006sparse} highlight that active selection causes non-smooth fluctuations in the marginal likelihood making the optimisation difficult. Hence, a different approach needs to be developed. 

\end{document}